\documentclass[conference]{IEEEtran}
\IEEEoverridecommandlockouts
\usepackage{cite}
\usepackage{amsmath,amssymb,amsfonts}
\usepackage{algorithm}
\usepackage{algorithmic}
\usepackage{graphicx}
\usepackage{subfig}
\usepackage{textcomp}
\usepackage{xcolor}

\def\BibTeX{{\rm B\kern-.05em{\sc i\kern-.025em b}\kern-.08em
    T\kern-.1667em\lower.7ex\hbox{E}\kern-.125emX}}
\begin{document}

\title{Cloud-based Federated Learning Framework for MRI Segmentation \\
\thanks{This work is supported by West Virginia University, Faculty Startup Grant (PI: A. El-Wakeel).}
}

\author{\IEEEauthorblockN{Rukesh Prajapati, and Amr S. El-Wakeel}
\IEEEauthorblockA{\textit{Lane Department of Computer Science and Electrical Engineering} \\
\textit{West Virginia University}\\
Morgantown, USA \\
rp000052@mix.wvu.edu, amr.elwakeel@mail.wvu.edu}
}

\maketitle

\begin{abstract}
    In contemporary rural healthcare settings, the principal challenge in diagnosing brain images is the scarcity of available data, given that most of the existing deep learning models demand extensive training data to optimize their performance, necessitating centralized processing methods that potentially compromise data privacy. This paper proposes a novel framework tailored for brain tissue segmentation in rural healthcare facilities. The framework employs a deep reinforcement learning (DRL) environment in tandem with a refinement model (RM) deployed locally at rural healthcare sites. The proposed DRL model has a reduced parameter count and practicality for implementation across distributed rural sites. To uphold data privacy and enhance model generalization without transgressing privacy constraints, we employ federated learning (FL) for cooperative model training. We demonstrate the efficacy of our approach by training the network with a limited data set and observing a substantial performance enhancement, mitigating inaccuracies and irregularities in segmentation across diverse sites. Remarkably, the DRL model attains an accuracy of up to 80\%, surpassing the capabilities of conventional convolutional neural networks when confronted with data insufficiency. Incorporating our RM results in an additional accuracy improvement of at least 10\%, while FL contributes to a further accuracy enhancement of up to 5\%. Collectively, the framework achieves an average 92\% accuracy rate within rural healthcare settings characterized by data constraints.
\end{abstract}


\section{Introduction}

Multiple prognostic assessments anticipate a nationwide deficiency in geriatricians and neurologists in the United States, with the most pronounced impact expected to be experienced within rural healthcare settings over the ensuing decades\cite{report_1} present, 20 states have been designated as 'dementia neurology deserts,' characterized by a projected availability of fewer than 10 neurologists per 10,000 residents by the year 2025\cite{report_2}. In addressing the insufficiency of specialized medical practitioners for diagnostic purposes in rural sites. However, the efficacy of deep learning hinges on the availability of relevant training data. It has been documented that diagnostic prevalence remains lower in rural counties compared to urban counterparts, and the persistent challenge of low patient volume in rural communities further complicates the application of deep learning methodologies\cite{report_3, report_4}. To address this issue, collecting data from rural healthcare sites at a central urban healthcare site can facilitate model training. However, the absence of data privacy would ensue if data were centralized within an urban location thereby contravening established data privacy protocols and concurrently augmenting expenditures attributable to the imperative need for physical transportation. 


One potential solution issue mentioned above issue involves the implementation of edge computing, wherein the machine learning model is trained locally at each rural healthcare institution using the data available at that specific site \cite{add_2}. However, it is essential to emphasize that achieving superior accuracy in model training necessitates access to a substantial volume of data. This resource may not be readily available at all healthcare institutions due to computational constraints. To address this challenge, an alternative approach involves the utilization of cloud computing, where the model is trained on a remote cloud server using data collected from various local sites. Cloud computing is utilized for applications and services over the Internet, along with the underlying software and hardware infrastructure that supports these services within data centers\cite{add_1}. Nevertheless, it is essential to acknowledge adoption of cloud computing introduces potential concerns related to data privacy and security, which warrant careful consideration in the context of healthcare applications.

\begin{figure*}
\centering
\includegraphics[width=6.5in, height=2.5in]{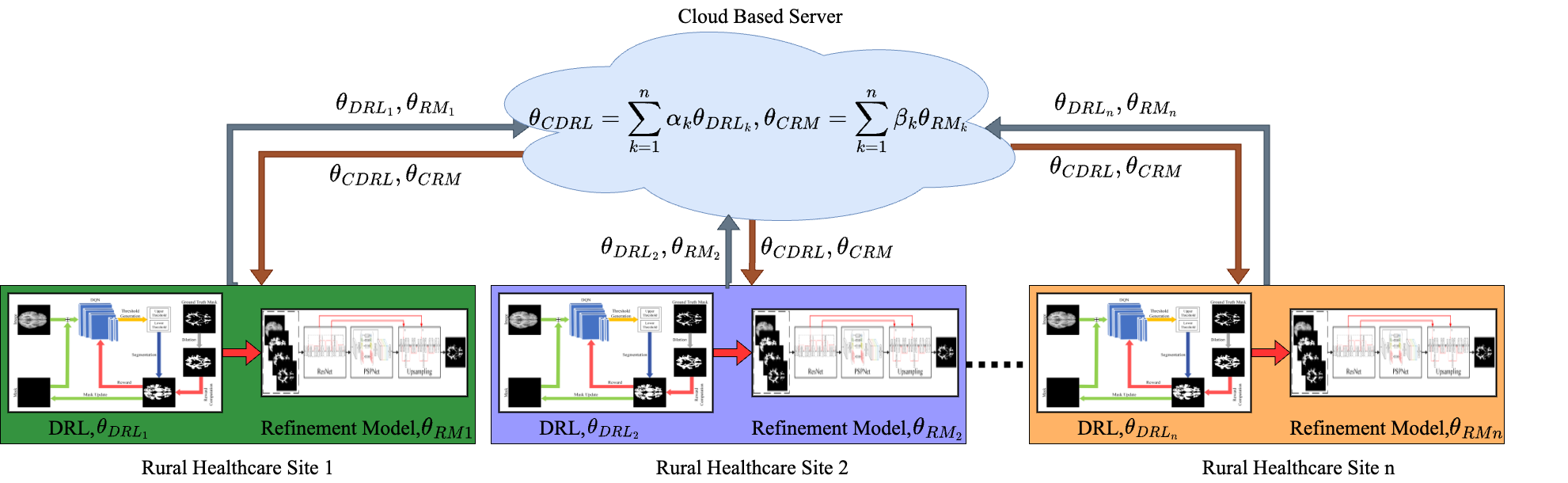}
\caption{Federated learning framework for MRI segmentation for rural healthcare sites.}
\label{fig}
\end{figure*}

For these issues currently afflicting rural hospitals in the United States, federated learning (FL) presents a potential remedy. The process of federated learning involves the periodic collection of locally trained models from individual healthcare sites or hospitals\cite{fed_1}. The objective is to derive shared global contributions from all participating sites\cite{fed_2}. After the aggregation of these local models, the resultant global model is broadcast back to the individual sites for ongoing training. This decentralized approach facilitates collaborative model training across multiple sites while safeguarding the privacy of local data, thus effectively addressing privacy concerns\cite{fed_3}. This collaborative multi-site model training approach holds the potential to enhance model generalization compared to single-site models. This improvement stems from the inherent diversity in the data from various rural healthcare institutions. Consequently, this approach can provide substantial benefits to rural sites that may have limited patient training data \cite{fed_4}.

To assist medical personnel, magnetic resonance imaging (MRI) is broadly applied for diagnostic and therapeutic purposes due to its information abundance and non-invasive technology approach. In MRI images, different modalities have been developed to capture specific anatomical characteristics. These modalities exhibit distinctions in terms of contrast and functionality \cite{article_1}. Three commonly referenced modalities are T1 (associated with spin-lattice relaxation), T2 (associated with spin-spin relaxation), and T2-flair (pertaining to fluid attenuation inversion recovery \cite{article_2}. T1 images are useful for examining structures in the brain, such as gray matter (GM) and white matter (WM) \cite{article_3}. GM or WM tissue atrophy is a well-known biomarker that helps diagnose neurodegenerative diseases like Alzheimer’s disease (AD) \cite{article_4}.

\begin{figure}[H]
\centerline{\includegraphics[width=85mm,scale=0.5]{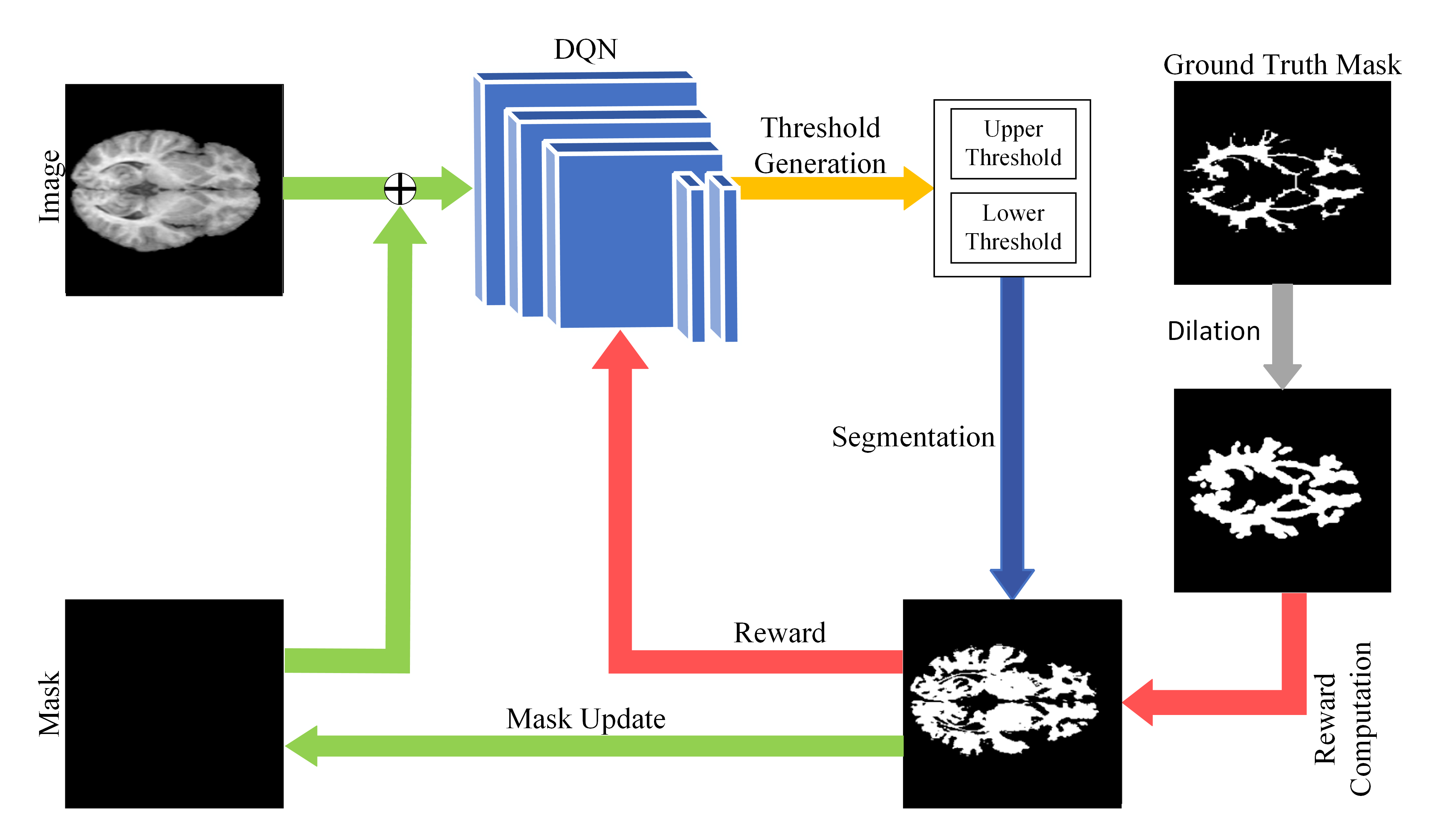}}
\caption{The DRL environment for the coarse segmentation.}
\label{fig2}
\end{figure}

Semantic segmentation does the crucial task of categorization and the identification of diseased tissues by assigning class labels to each pixel in an MRI image. Many algorithms and techniques have been developed so far for tissue segmentation\cite{article_7}. Deep learning has recently made significant strides in medical image segmentation, with successful outcomes observed in both 2D and 3D convolutional neural networks (CNNs) for MRI image segmentation \cite{article_15,article_16}. To achieve higher output accuracy, these models need to be trained on large data sets. Unfortunately, for individual institutions, the acquisition and processing of large data sets can be challenging due to cost constraints, computational complexity, and patient medical records privacy.


In this article, we present an innovative MRI segmentation architecture for local healthcare sites using FL and DRL approaches. Unlike previous deep learning models, our proposed model can achieve improved results for rural healthcare sites even when trained with a limited number of subjects. Our approach involves a two-stage segmentation model comprising coarse segmentation and refinement model. The DRL environment enables segmentation and coarse segmentation generation with a reduced number of model parameters. DRL actions are defined as threshold levels that generate thresholds for the input MRI image, addressing the overlap between the intensity levels of WM and GM to produce coarse segmentation. The refinement network is capable of adjusting both local and global boundaries and then enhances the edges of the DRL output.

\section{Framework and Methodology}

\subsection{Dataset}
In our study, we assess the performance of our proposed DRL model using the Internet Brain Segmentation Repository (IBSR), which consists of data from 18 subjects. Employing a random sampling approach, we select five subjects from each healthcare site represented in the repository. Subsequently, we extract 2D images from each subject along three distinct anatomical planes: axial, coronal, and sagittal. Our research primarily focuses on the segmentation of a WM. 
Our objective is to train a model effectively with a limited data set, motivating our choice of the IBSR data set, which contains MRI images from only 18 subjects.

\subsection{Proposed Framework}
\subsubsection{Federated Learning}
The proposed framework for FL-based segmentation delineates a network of multiple rural healthcare institutions interconnected through a central cloud-based server, as illustrated in Fig.~\ref{fig}. Within this architecture, each rural healthcare facility possesses its individual DRL environment for generating initial coarse segmentation, along with a Refinement Module (RM) designed to enhance these preliminary results. Initially, each rural site's model is independently trained on-site, and subsequently, the parameters associated with both the DRL and RM components are transmitted to the central cloud-based server. The cloud server undertakes parameters of local DRL $\theta_{DRL_1}, \theta_{DRL_2}... \theta_{DRL_n}$ and parameters of refinement model $\theta_{RM_1}, \theta_{RM_2}... \theta_{RM_n}$. The aggregated parameters $\theta_{CDRL}$ and $\theta_{CRM}$ are broadcast back to all rural healthcare sites.
\begin{equation}
\theta_{CDRL} = \sum_{k=1}^{n} \alpha_{k} \theta_{DRL_{k}}
\end{equation}

\begin{equation}
\theta_{CRM} = \sum_{k=1}^{n} \beta_{k} \theta_{RM_{k}}
\end{equation}
where $\theta_{CDRL}$ and $\theta_{CRM}$ are aggregated parameters for cloud-based DRL and RM models, respectively. $\alpha_{k}$ and $\beta_{k}$ are set to $\frac{N_{DRL_{k}}}{N_{DRL}}$ and $\frac{N_{RM_{k}}}{N_{RM}}$ where $N_{DRL_{k}}$, and $N_{RM_{k}}$ denotes number of training samples for DRL and RM at $k^{th}$ site respectively. $N_{DRL}$ is total DRL training images across all healthcare sites and $N_{RM}$ is total RM training images across all healthcare sites.

For subsequent rounds, the broadcasted parameters are once again employed for local training at each rural site and subsequently forwarded to the central cloud-based server. This iterative process continues for several rounds, ultimately yielding the final parameters utilized across all participating local sites. It is noteworthy that, in this framework, only the model parameters are shared with the cloud server, whereas the actual data from each local site remains unshared. This approach is devised to uphold the stringent data privacy policies enforced by institutional sites. Moreover, the models employed at the rural sites are characterized by their compact size, resulting in a reduced parameter footprint in contrast to the conventional deep learning models. This attribute renders the framework amenable to edge computing environments.

It should be emphasized that the data set has been partitioned in such a manner that certain rural institutional sites are furnished with significantly fewer training data points compared to the quantities typically employed in training deep learning models. The goal of this framework is to achieve heightened accuracy, even when dealing with sites possessing a limited volume of patient data. This emphasis on performance improvement is particularly pertinent given the primary focus of the framework on rural healthcare institutions.

\subsubsection{ThreshNet}

We introduce ThreshNet, a novel automated threshold generation system designed for threshold segmentation tasks. ThreshNet's core component is a DRL agent as shown in Fig.~\ref{fig2}, complemented by an off-the-shelf threshold segmentation process. The system operates as follows: an MRI image and a blank mask image are concatenated and input into the DQN model. The DQN then suggests new threshold levels based on these inputs (Algorithm \ref{alg:threshold}), which are used to compute upper and lower threshold values. Subsequently, a threshold-based segmentation is applied to generate a new binary mask. This binary mask serves a dual purpose: it is used to compute rewards by comparing it to the ground truth mask and also serves as an observation for the subsequent iteration. During training, ThreshNet iteratively performs this cycle, while during testing, only the threshold generation procedure is executed. The final segmentation masks are obtained by repeating the threshold generation process three times for each input MRI image and performing threshold segmentation accordingly.

\begin{algorithm}
\caption{Threshold calculation}\label{alg:threshold}
\textbf{Input:} Maximum pixel value of an input image $max_{p}$, upper threshold level $level_{upper}$, and lower threshold level $level_{lower}$\\
\textbf{Output:} upper threshold $th_{upper}$  and  lower threshold $th_{lower}$
\begin{algorithmic}[1]
\STATE $max = max_{p}, min = \frac{max_{p}}{2}$
\STATE $v_{level} = \frac{max-min}{50}$
\STATE $th_{upper} = min + v_{level}*level_{upper}$
\STATE $th_{lower} = min + v_{level}*level_{lower}$
\end{algorithmic}
\end{algorithm}

\begin{figure}[htbp]
\centerline{\includegraphics[width=85mm,scale=0.5]{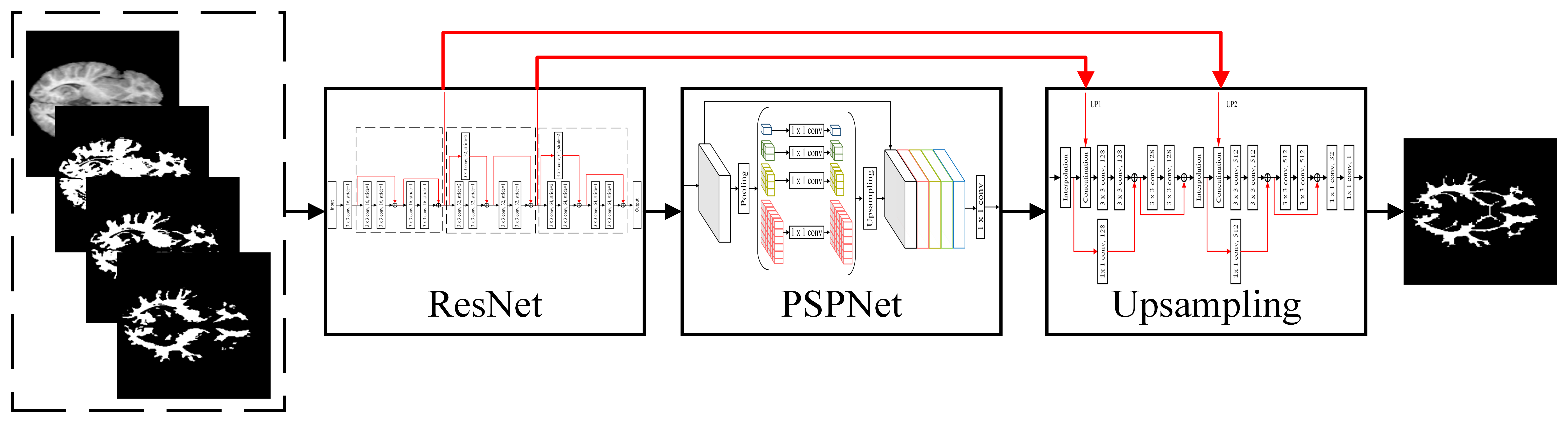}}
\caption{Network architecture of refinement model based on cascadePSP with reduced parameters.}
\label{fig3}
\end{figure}

\begin{figure*}[t]
\centering
\begin{tabular}{cccccc}
\subfloat[Input]{\includegraphics[width = 1in]{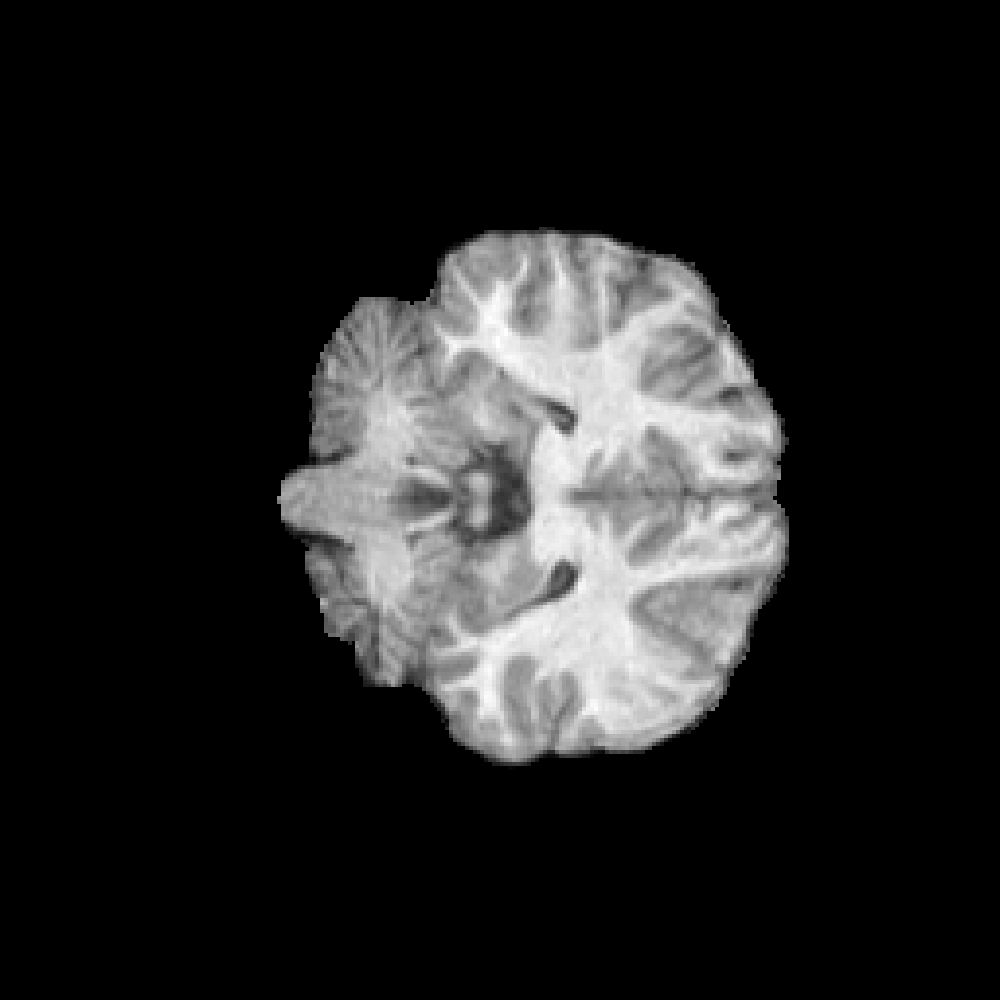}} &
\subfloat[Groudn truth]{\includegraphics[width = 1in]{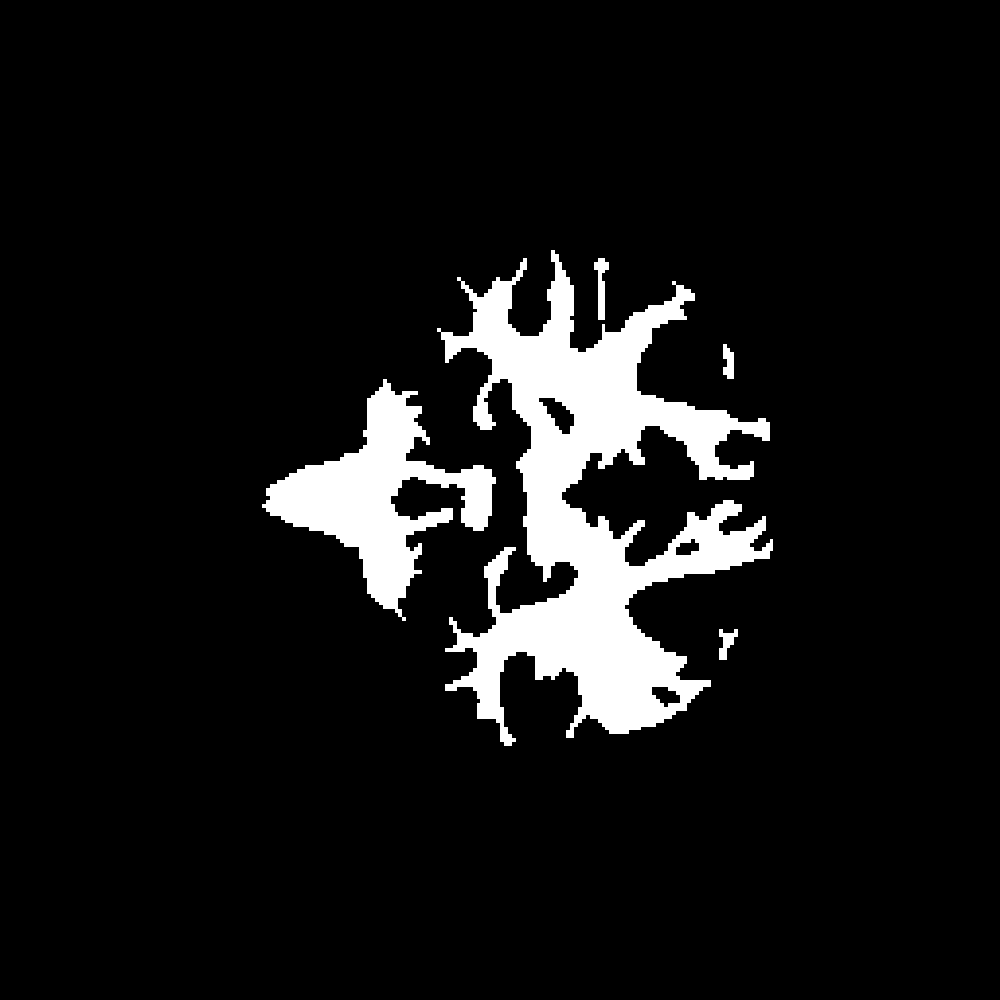}} &
\subfloat[DRL round 1]{\includegraphics[width = 1in]{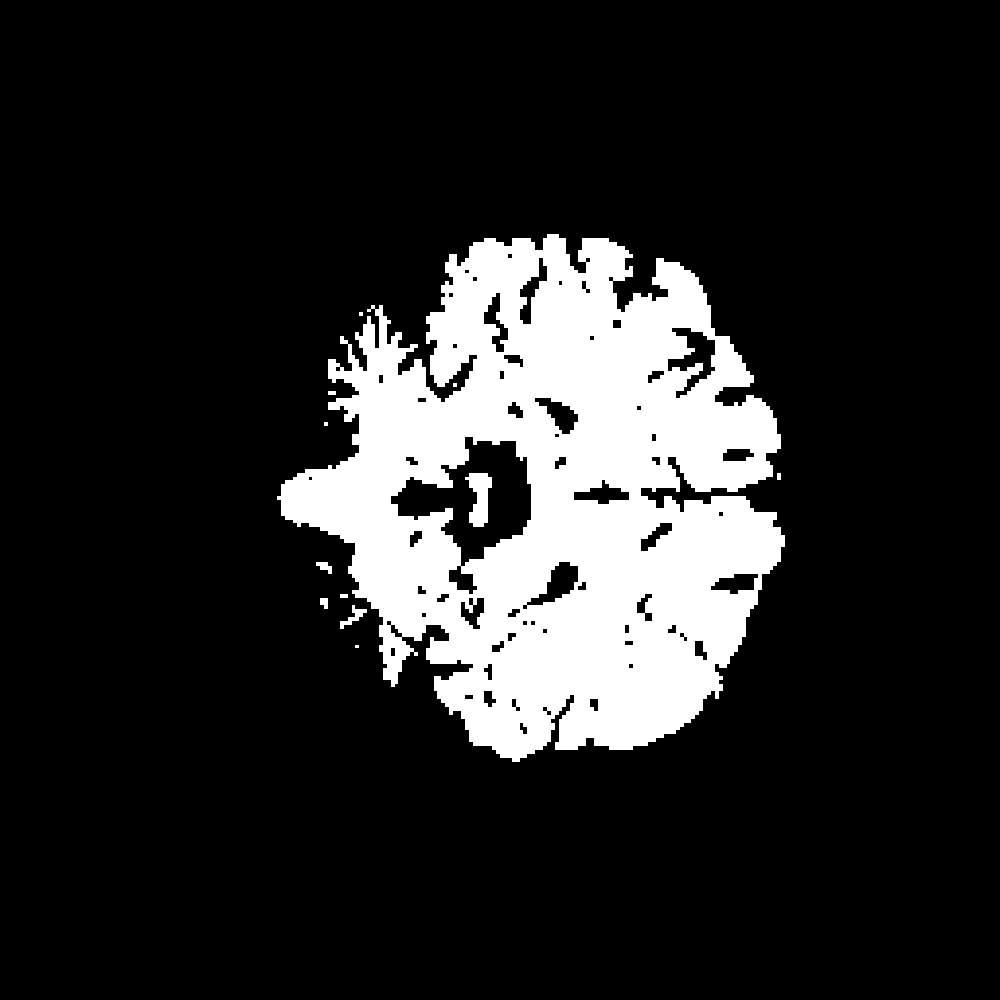}} &
\subfloat[DRL round 2]{\includegraphics[width = 1in]{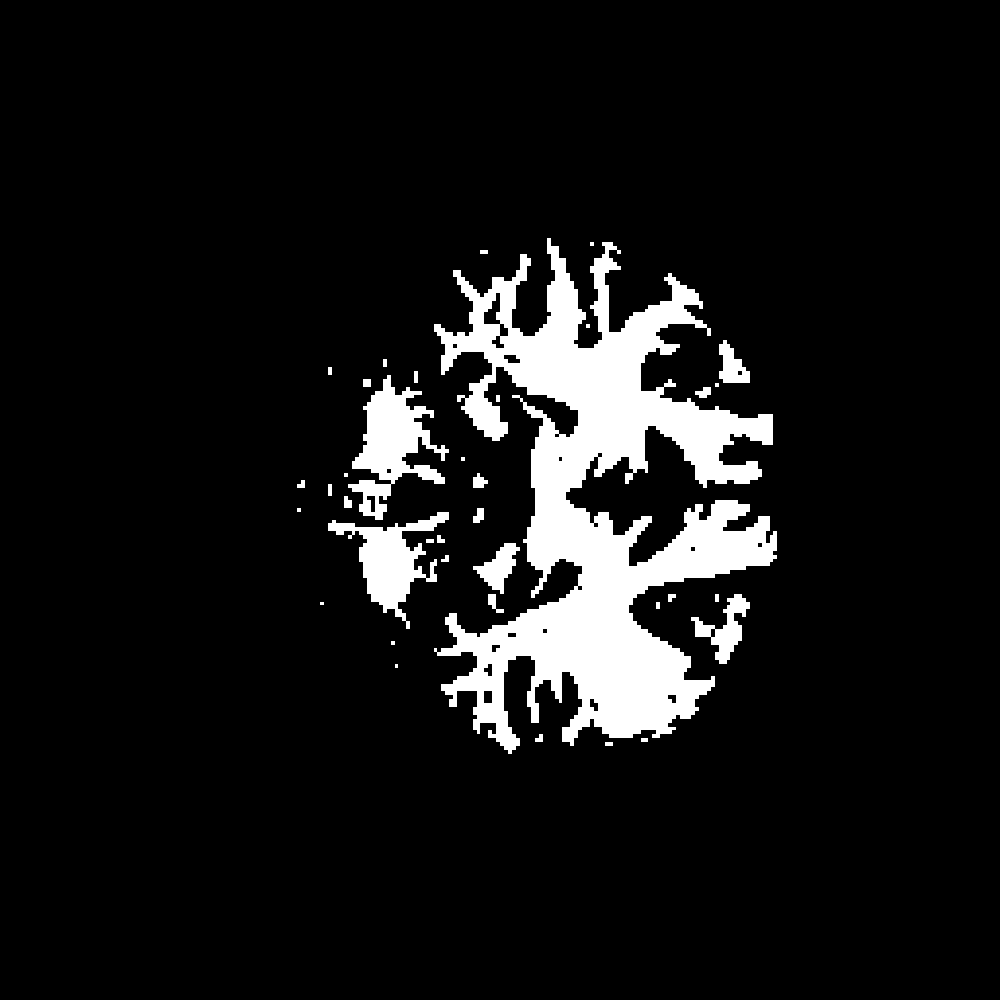}}&
\subfloat[DRL round 3]{\includegraphics[width = 1in]{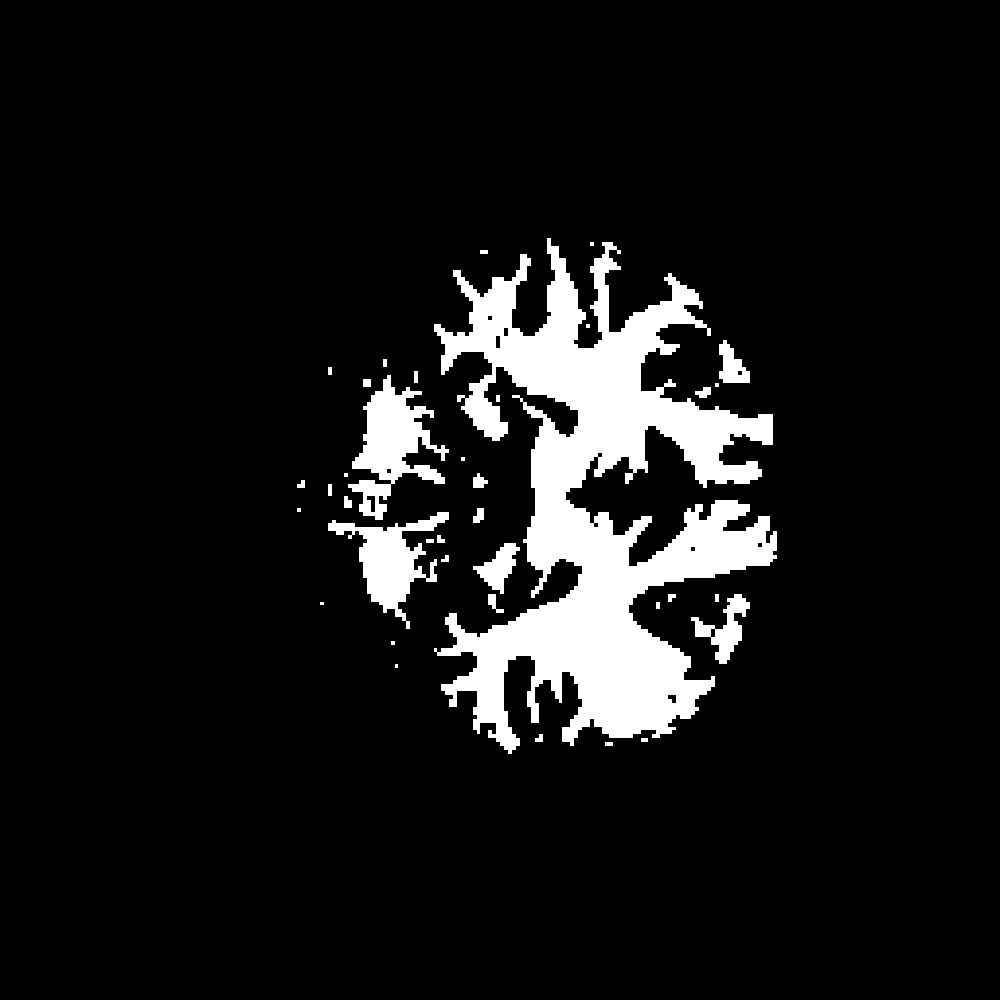}}&
\subfloat[Final (RM) output]{\includegraphics[width = 1in]{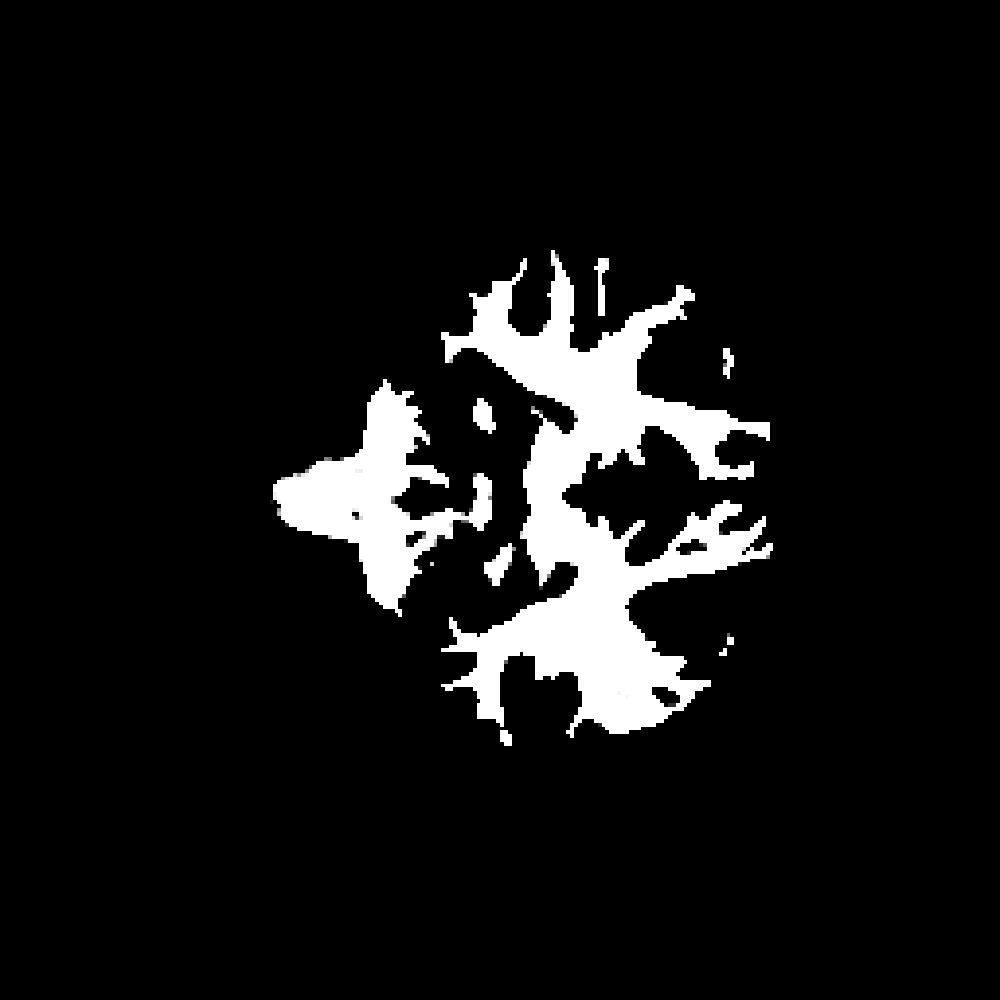}}\\
\setcounter{subfigure}{0}
\subfloat[Input]{\includegraphics[width = 1in]{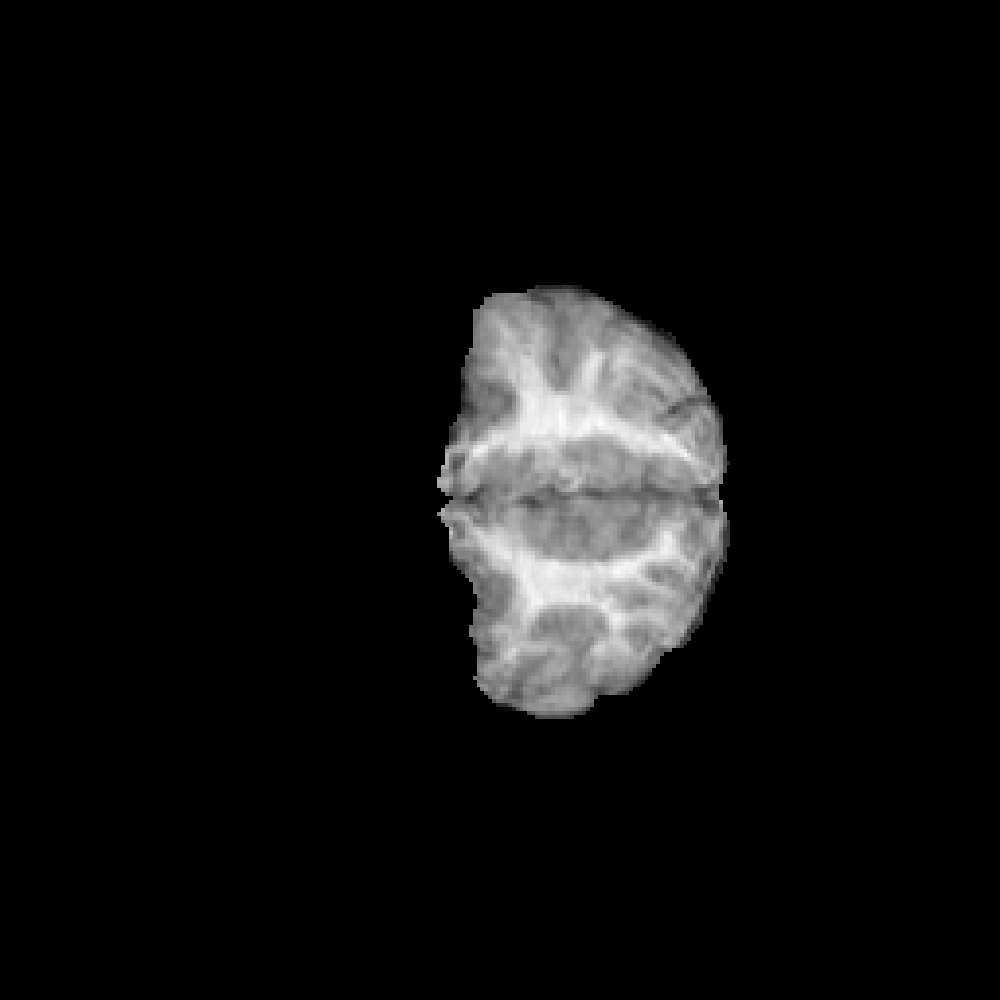}} &
\subfloat[Ground truth]{\includegraphics[width = 1in]{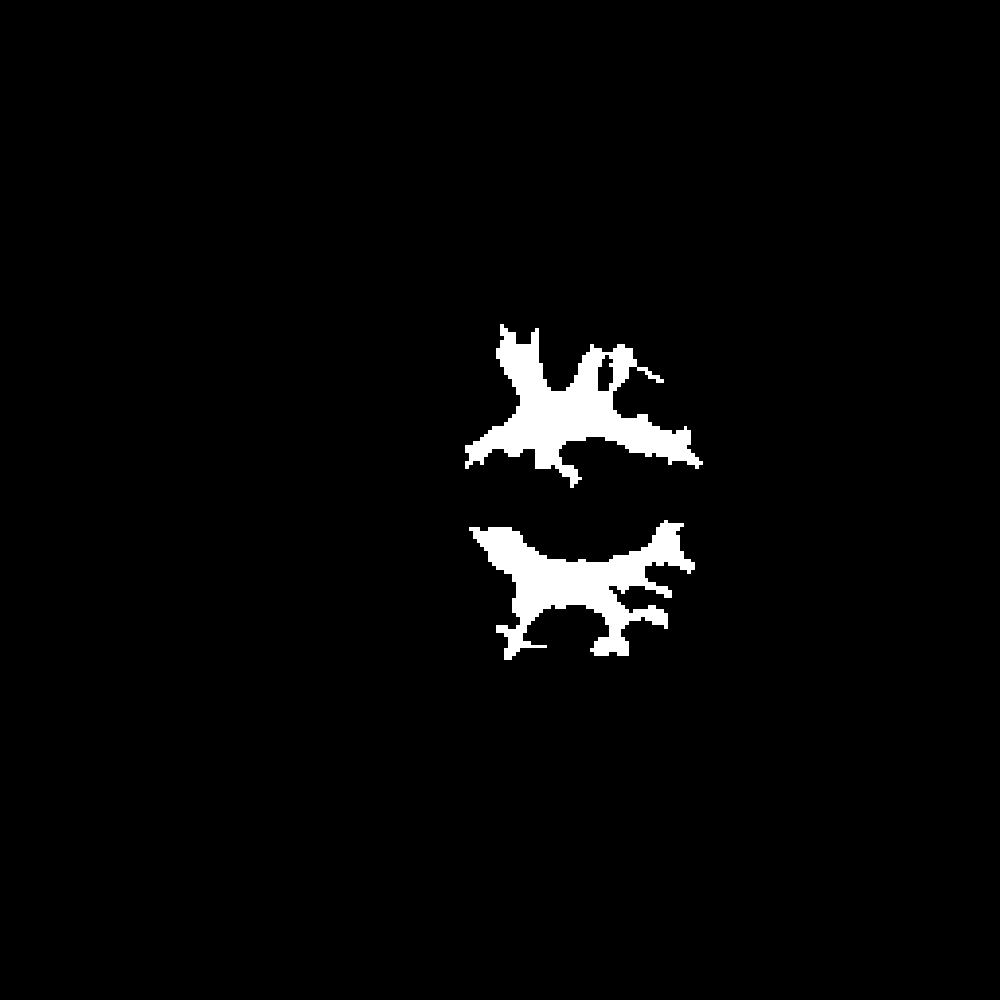}} &
\subfloat[DRL round 1]{\includegraphics[width = 1in]{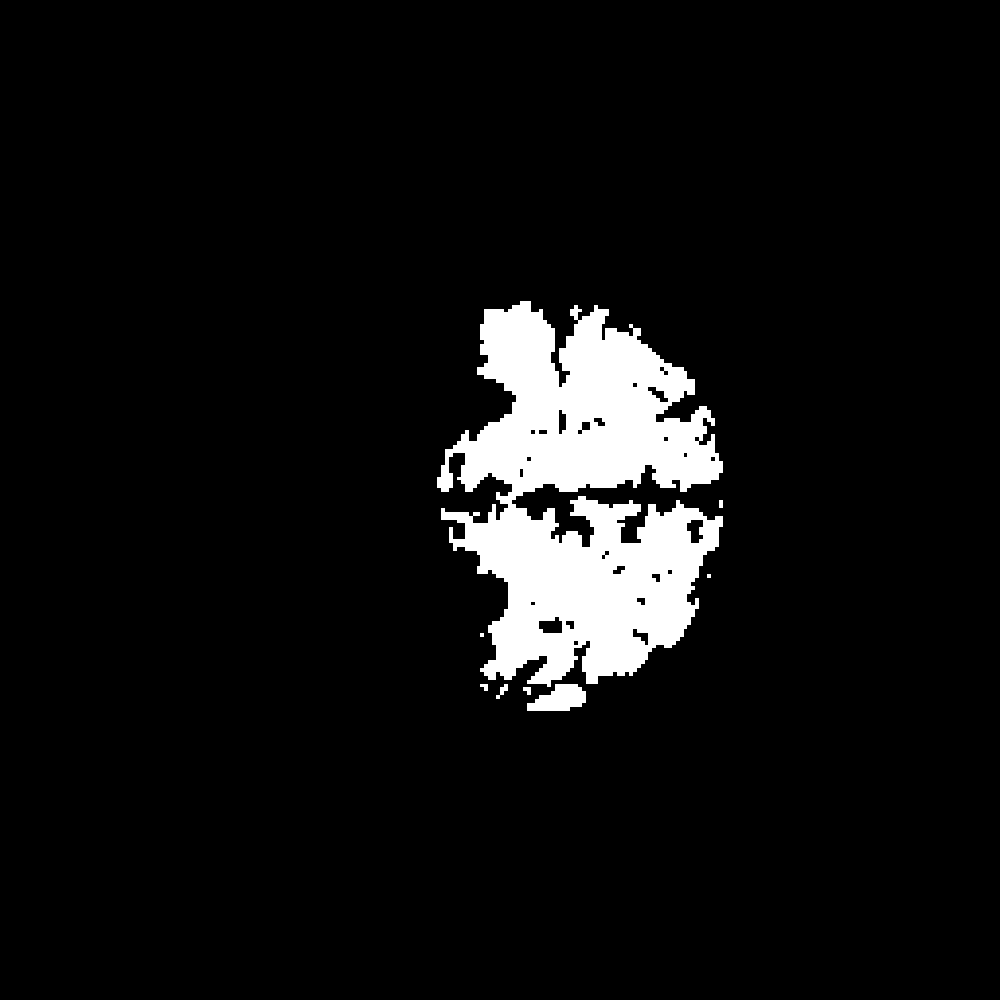}} &
\subfloat[DRL round 2]{\includegraphics[width = 1in]{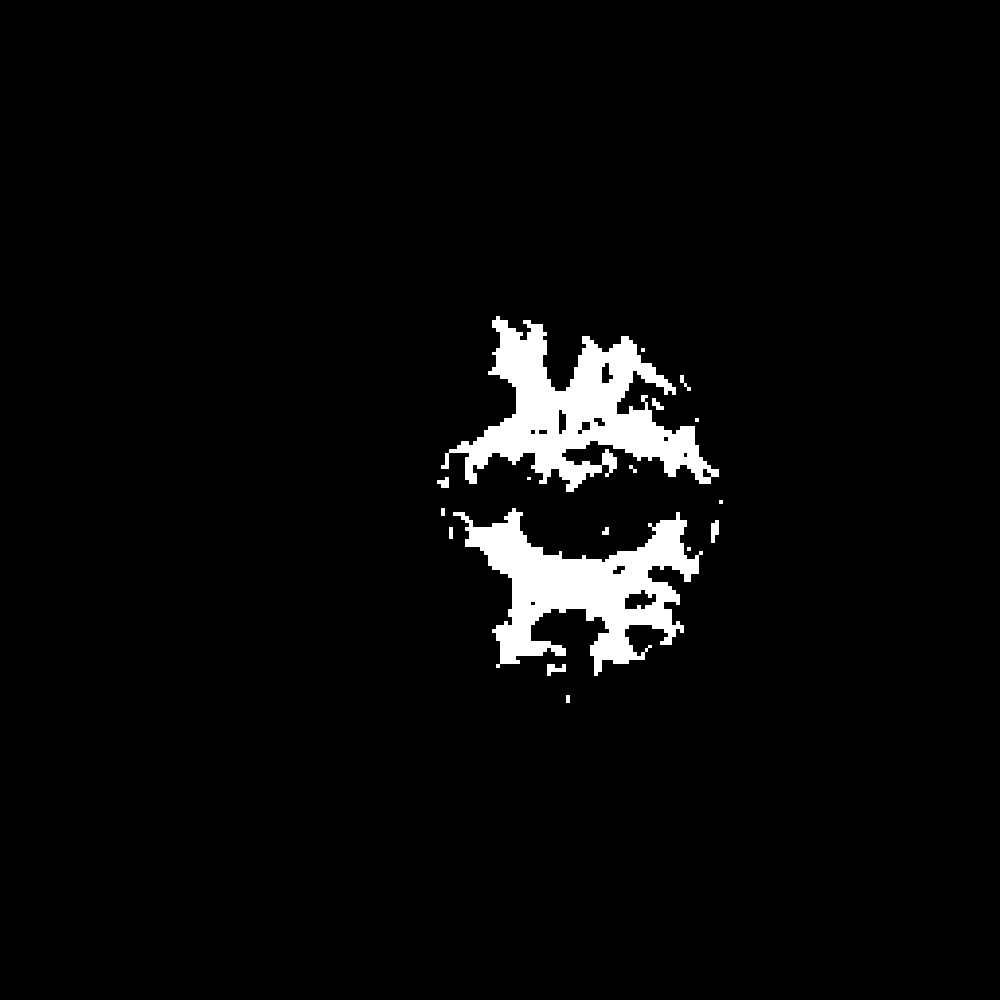}}&
\subfloat[DRL round 3]{\includegraphics[width = 1in]{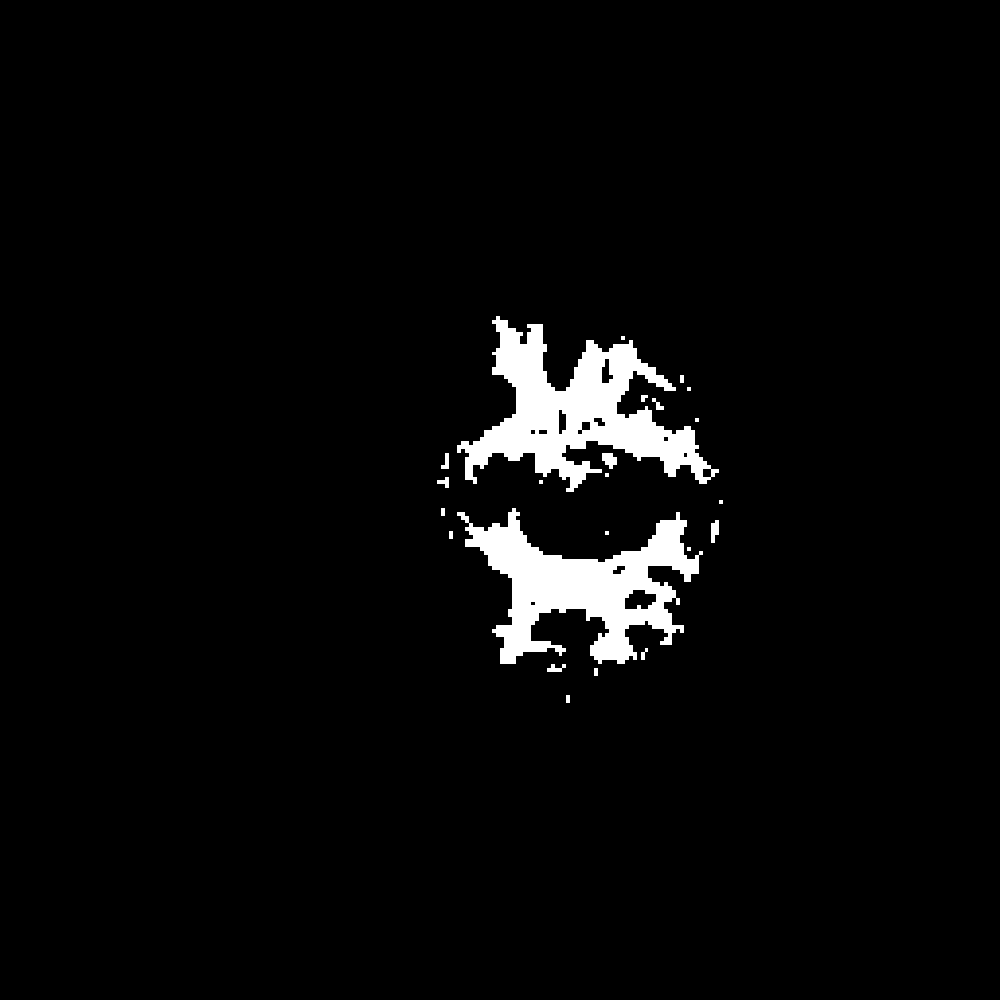}}&
\subfloat[Final (RM) output]{\includegraphics[width = 1in]{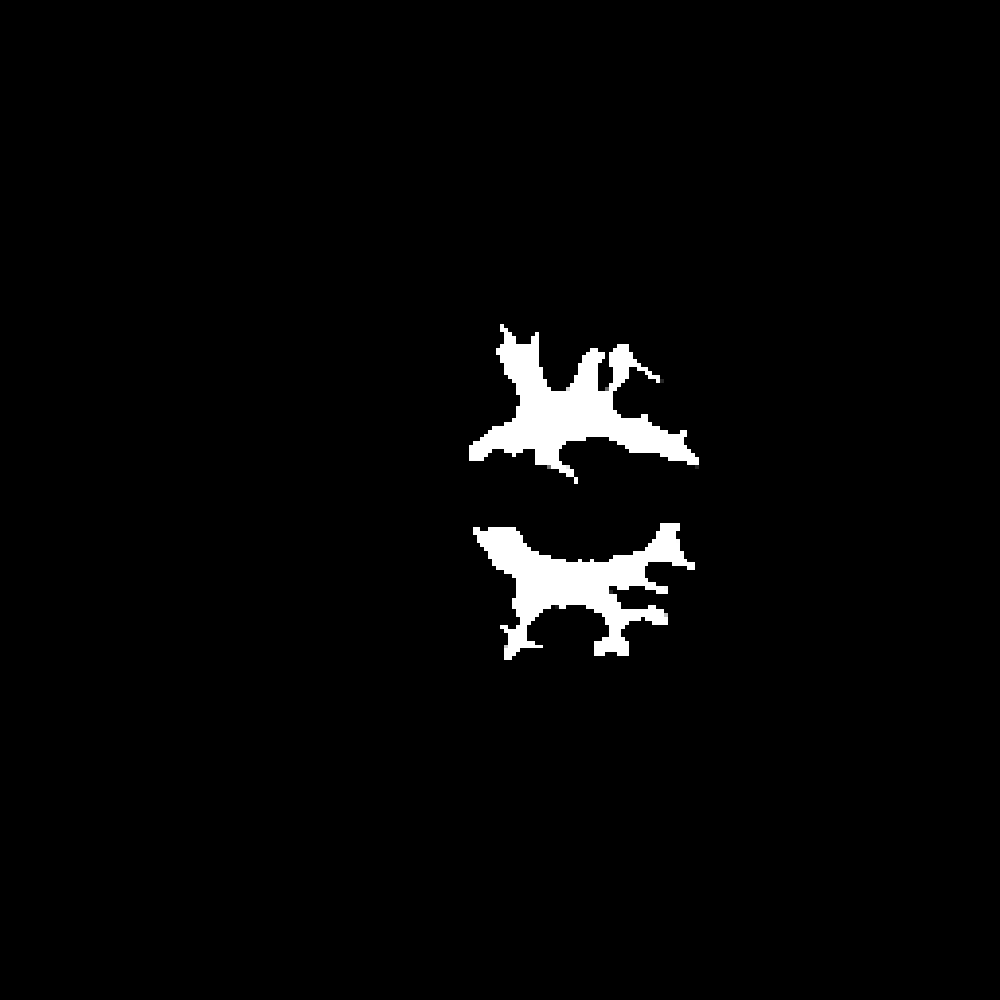}}\\
\setcounter{subfigure}{0}
\subfloat[Input]{\includegraphics[width = 1in]{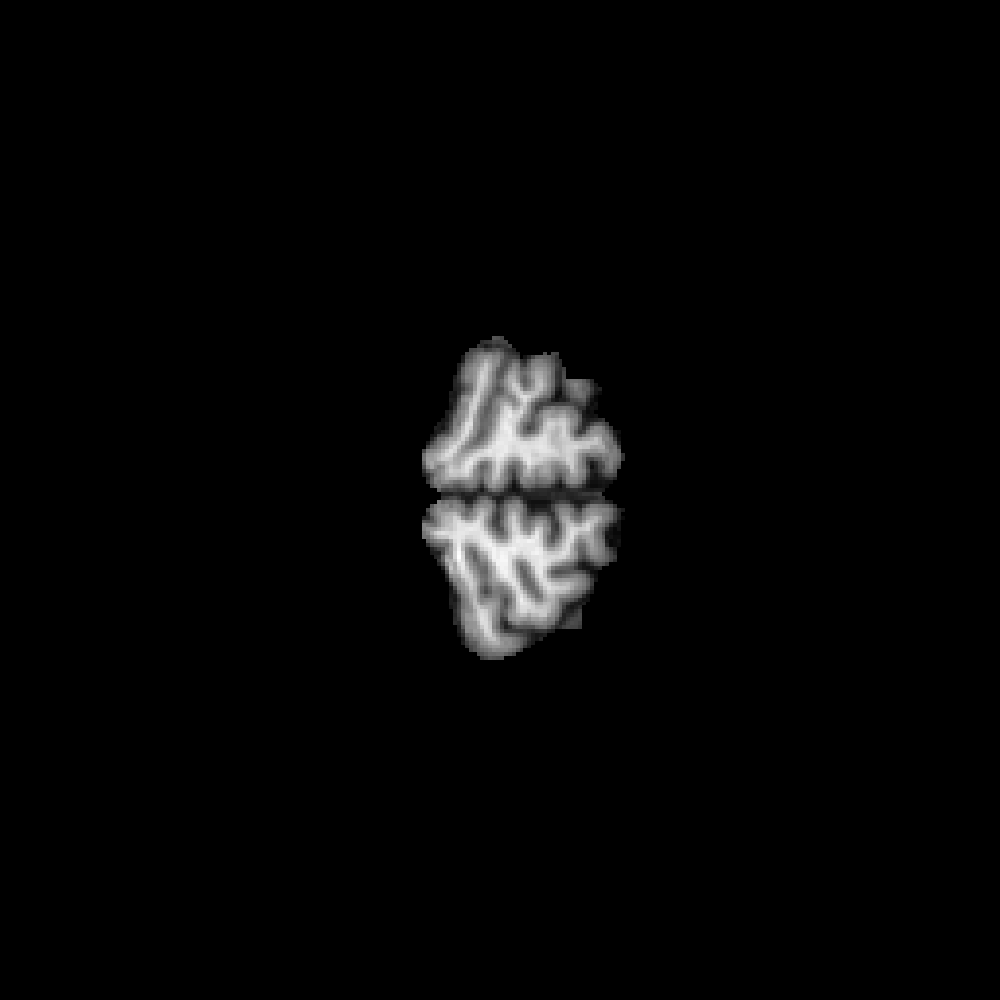}} &
\subfloat[Ground truth]{\includegraphics[width = 1in]{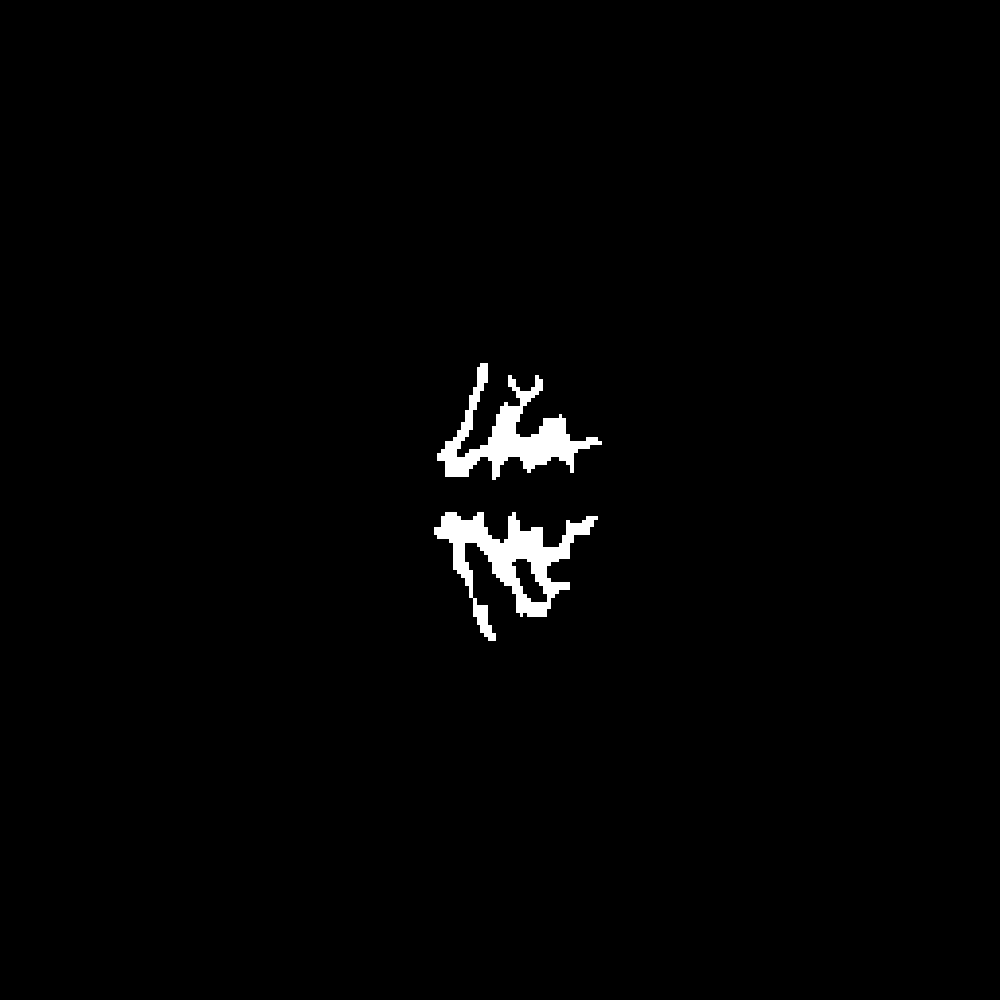}} &
\subfloat[DRL round 1]{\includegraphics[width = 1in]{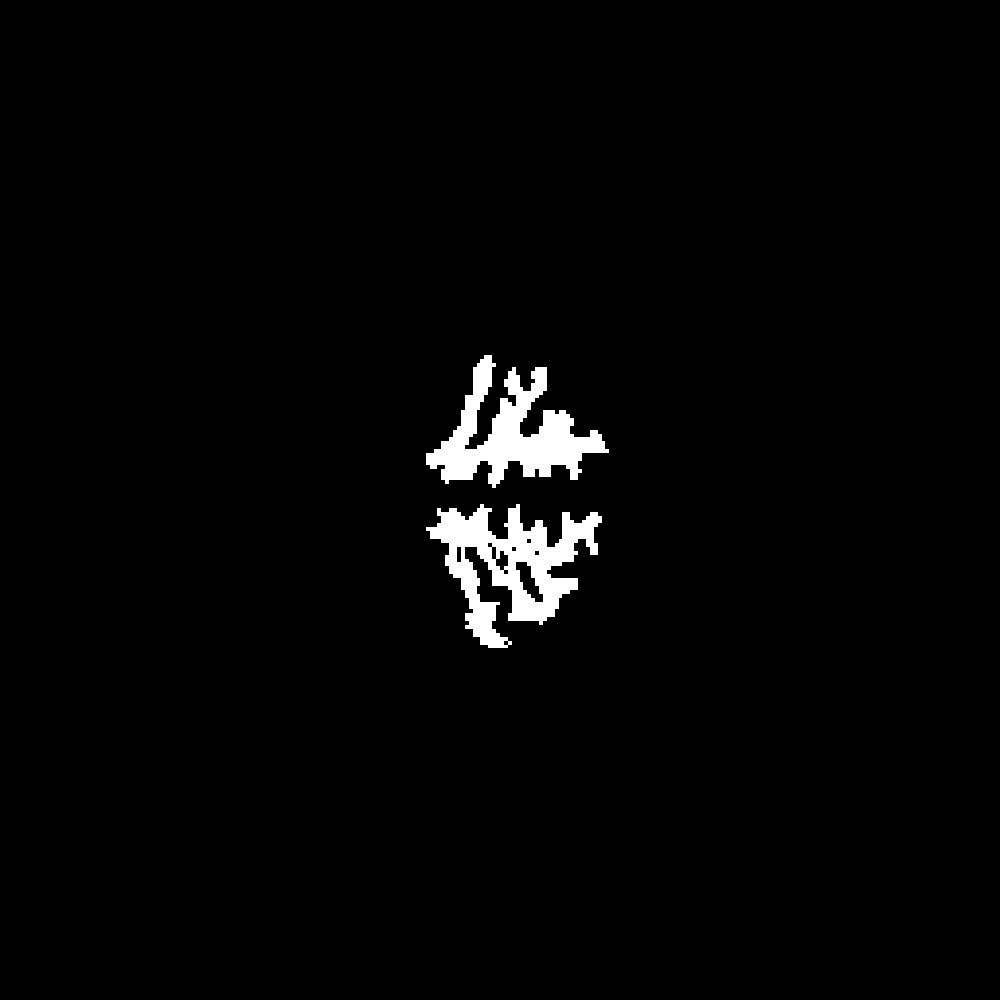}} &
\subfloat[DRL round 2]{\includegraphics[width = 1in]{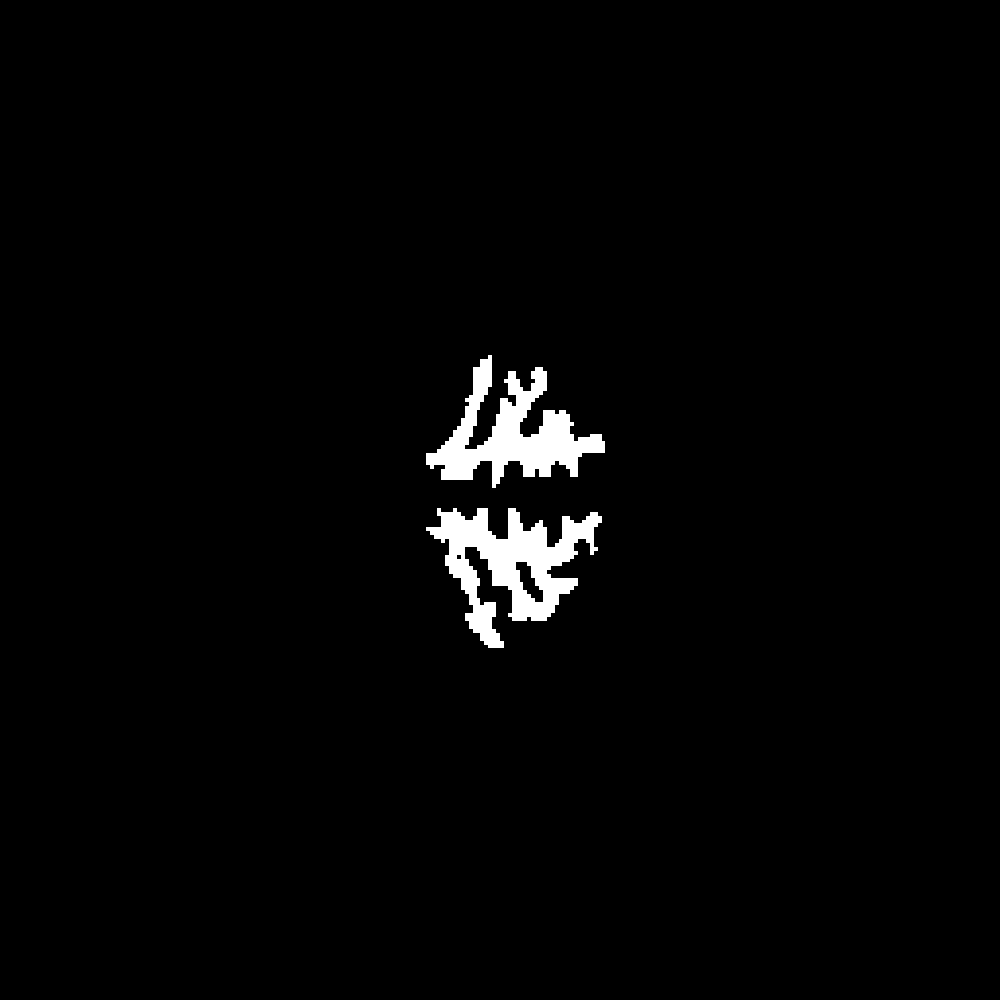}}&
\subfloat[DRL round 3]{\includegraphics[width = 1in]{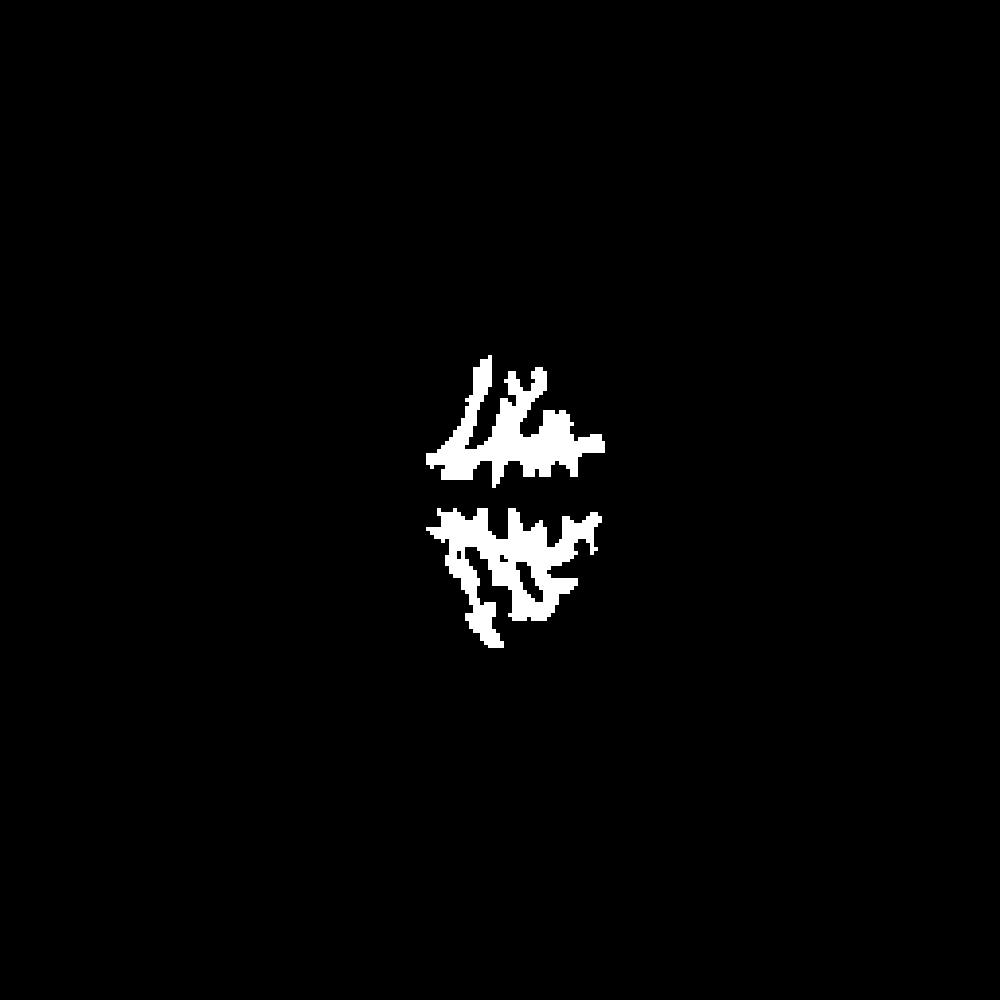}}&
\subfloat[Final (RM) output]{\includegraphics[width = 1in]{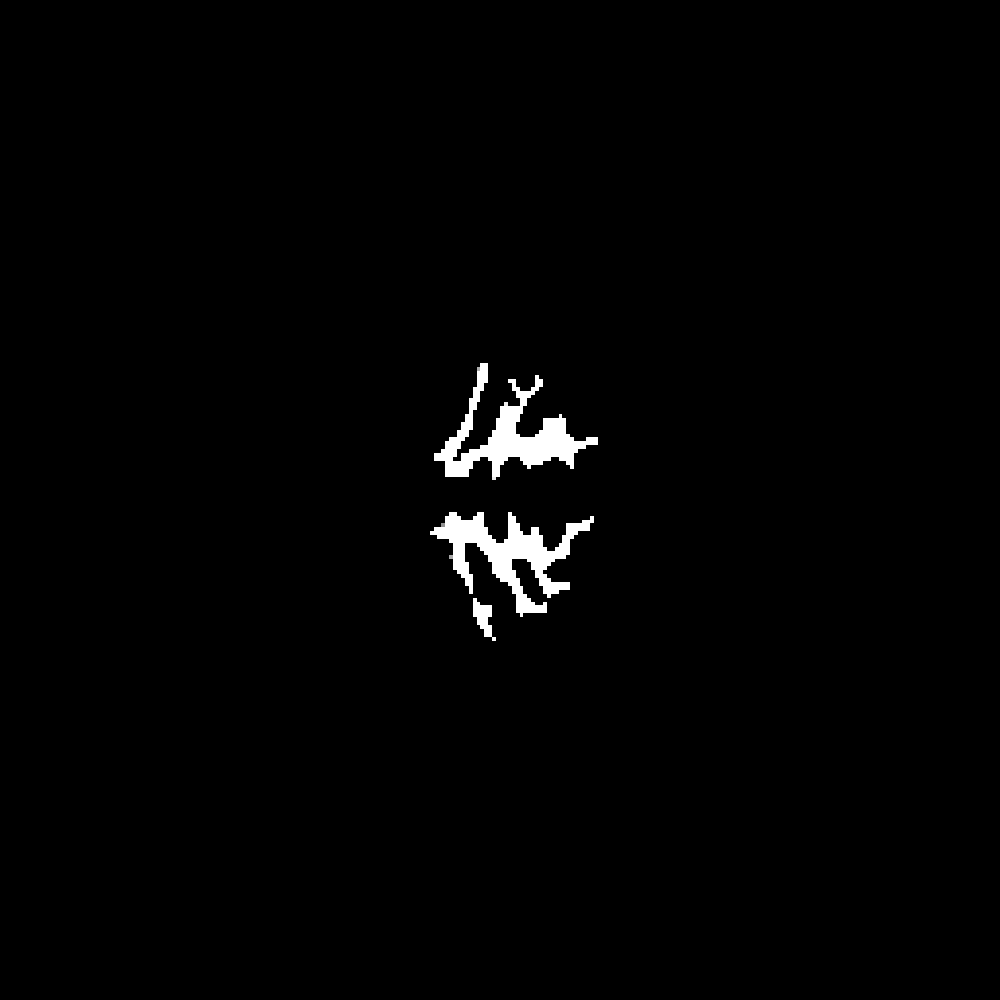}}
\end{tabular}
\caption{Input MRI image and their corresponding output images from models, where the first row shows the outputs from rural healthcare site 1, the second row shows from site 2, and the third row shows from site 3}
\label{fig4}
\end{figure*}

We delineate the problem by formalizing it as a Markov Decision Process (MDP) comprising three integral components: state, action, and reward.\\
\textbf{State:} The state should possess a requisite degree of intricacy to facilitate the agent's informed decision-making process. In the case of ThreshNet, the state is formulated by concatenating the MRI image with a segmented mask at each discrete step.\\
\textbf{Action:} The agent selects an action from the action space, defined as a new threshold level based on the current state. To manage the challenges associated with training, we limit the threshold levels' grid to a sparse 50 $\times$ 2 size, resulting in a total of 100 possible actions encompassing both upper and lower threshold adjustments. In each step, the agent chooses two actions, leading to the calculation of new threshold values at their corresponding levels. It is important to note that due to the absence of a clearly defined terminal action, the process concludes after recommending three sets of threshold values.\\
\textbf{Reward:} The reward signal evaluates the agent's actions, with mask precision relative to the ground truth serving as a scoring criterion. While the Dice Similarity Coefficient (DSC) is the straightforward choice for a reward function, we also incorporate the exponential DSC model to accentuate improvements in higher DSC levels, as described in \cite{add_3}.
\begin{equation}
    R_{exp}=\frac{exp^{k*DSC(M,G_{d})}-1}{exp^{k}-1}
\end{equation}
where $M$ denotes the segmented mask, $G_{d}$ denotes the dilated ground truth, and $k$ denotes a constant value. In ThreshNet, success or failure is determined based on the DSC score of the newly segmented mask. We establish a reward condition: if the segmented mask exhibits over 70\% DSC similarity, we apply the exponential DSC reward; otherwise, a reduced reward signal is employed.
\begin{equation}
    R=
    \begin{cases}
    R_{exp}, & \text{if}\  DSC > 0.7\\
    R_{exp} - 1, & \text{otherwise}
    \end{cases}
\end{equation}

\subsubsection{Refinement Model}
In the pursuit of enhancing the refinement process, we adopted the cascadePSP\cite{article_24}  architecture, which exhibits the capacity to refine and rectify local boundaries whenever feasible. This architectural framework is distinguished by its adaptability, rendering it applicable across a spectrum of resolutions without necessitating the intricate task of fine-tuning its parameters. Furthermore, to render the cascadePSP architecture conducive to edge computing within the confines of rural healthcare facilities, we undertook a parameter reduction initiative. In the original architecture, the ResNet-50 module was employed; however, as illustrated in Fig. \ref{fig3}, we substituted it with three blocks of residual modules.

\begin{table}[H]
\caption{Comparison of parameter numbers}
\begin{center}
    \begin{tabular}{c|c}
         \textbf{Models} & \textbf{Number of parameters in million}  \\
         \hline
         UNet \cite{unet} &  31\\
         UNet++ \cite{unet++} & 36.6\\
         R2UNet \cite{r2unet} & 101.9\\
         Proposed (DRL + RM) & (5.4 + 8.6) = 13
    \end{tabular}
\label{comp}
\end{center}
\end{table}

The cumulative count of parameters in the fusion of both DRL and RM approaches approximates 13 million. Table \ref{comp} presents a comparative analysis of these parameters regarding various established deep learning models. Remarkably, the parameter count in our proposed model is nearly threefold less than that of the baseline UNet model \cite{unet}, yet it demonstrates superior performance. This parameter reduction makes our proposed model amenable for implementation in resource-constrained rural settings, where advanced computational resources may be lacking.

\section{Results and Discussions}

\begin{table*} 
\small
\hfill
\hfill
\caption{Segmentation results of the proposed framework at each rural healthcare site without FL}
\begin{center}
\begin {tabular}{|c|c|c|c|c|c|c|}
\hline
\multicolumn{7}{|c|}{\textbf{Experiment 1 with data distribution of 10:4:4 ratio}}\\
\hline
\multicolumn{4}{|c|}{\textbf{Output result from DRL model}} & \multicolumn{3}{|c|}{\textbf{Output result from DRL model + RM }} \\
\hline
\textbf{Scores} & \textbf{Site 1}& \textbf{Site 2}& \textbf{Site 3} & \textbf{Site 1}& \textbf{Site 2}& \textbf{Site 3}\\
\hline
DSC &0.7663$\pm$0.1519 &  0.7974$\pm$0.1502 &  0.7918$\pm$ 0.1225 &0.9425 $\pm$ 0.0233 & 0.8909 $\pm$ 0.1228 & 0.9335 $\pm$ 0.0738\\

Sensitivity & 0.9348$\pm$0.1409  &  0.8985$\pm$0.1425 &  0.9546$\pm$0.0521 &0.9329 $\pm$ 0.0391 & 0.9107 $\pm$ 0.0533 & 0.9484 $\pm$ 0.0297\\

Specificity & 0.9722$\pm$0.0258  &  0.9833$\pm$0.0147 &  0.9753$\pm$0.0177 &0.9972 $\pm$ 0.0028 & 0.9939 $\pm$ 0.053 & 0.9968 $\pm$ 0.0022\\

MAE & 0.03034$\pm$ 0.0245 &  0.0207$\pm$0.0147 &  0.0264$\pm$0.018 &0.0059 $\pm$ 0.0037 & 0.0091 $\pm$ 0.0064 & 0.0061 $\pm$ 0.0039\\
\hline
\multicolumn{7}{|c|}{\textbf{Experiment 2 with data distribution of 6:6:6 ratio}}\\
\hline
DSC & 0.7762$\pm$0.1274  &  0.7704$\pm$0.18 &  0.6625$\pm$0.2201 &0.929 $\pm$ 0.0885 & 0.9373 $\pm$ 0.0303 & 0.8931 $\pm$ 0.1229\\

Sensitivity &0.9352$\pm$0.1125  &  0.8908$\pm$0.1877 &  0.9006$\pm$0.2112 &0.9096 $\pm$ 0.048 & 0.9421 $\pm$ 0.0377 & 0.9095 $\pm$ 0.0786\\

Specificity &0.9735$\pm$0.0209  &  0.9749$\pm$0.0225 &  0.962$\pm$0.0347  &0.9977 $\pm$ 0.002 & 0.9956 $\pm$ 0.0026 & 0.9951 $\pm$ 0.0032\\

MAE &0.0282$\pm$0.0199  &  0.0313$\pm$0.0224 &  0.0403$\pm$0.0336 &0.0066 $\pm$ 0.0042 & 0.0073 $\pm$ 0.0037 & 0.0083 $\pm$ 0.006\\
\hline
\end{tabular}
\label{tab1}
\end{center}
\end{table*}

\begin{table}[t]
\caption{Segmentation results of the proposed framework at each rural healthcare site with FL}
\begin{center}
\begin{tabular}{|c|c|c|c|}
\hline
\multicolumn{4}{|c|}{\textbf{DRL model and RM trained with FL}}\\
\hline
\multicolumn{4}{|c|}{\textbf{Experiment 1 with data distribution of 10:4:4 ratio}}\\

\hline
\textbf{Scores} & \textbf{Site 1}& \textbf{Site 2}& \textbf{Site 3}\\
\hline
DSC & 0.9345 $\pm$ 0.0331 & 0.9247 $\pm$ 0.1233 & 0.9515 $\pm$ 0.0399\\

Sensitivity &0.9597 $\pm$ 0.0346 & 0.9596 $\pm$ 0.0339 & 0.9579 $\pm$ 0.0293\\

Specificity &0.9952 $\pm$ 0.0042 & 0.9961 $\pm$ 0.0032 & 0.9972 $\pm$ 0.0023\\

MAE &0.068 $\pm$ 0.0045 & 0.0055 $\pm$ 0.0033 & 0.0049 $\pm$ 0.0035\\
\hline
\multicolumn{4}{|c|}{\textbf{Experiment 2 with data distribution of 6:6:6 ratio}}\\
\hline
DSC &0.9424$\pm$0.032  &  0.9424$\pm$0.0347  &  0.9361$\pm$0.1049 \\

Sensitivity & 0.9752$\pm$0.0386  &  0.9566$\pm$0.0345  &  0.9758$\pm$0.0334 \\

Specificity & 0.9947$\pm$0.0041  &  0.9959$\pm$0.003  &  0.9961$\pm$0.0031 \\

MAE & 0.0062$\pm$0.0043  &  0.0065$\pm$0.0033  &  0.0048$\pm$0.0035 \\
\hline
\end{tabular}
\label{tab2}
\end{center}
\end{table}

We developed a system that encompasses three rural healthcare sites. The analyzed subjects were distributed among these rural sites. Our primary objective is to enhance the accuracy of these rural sites while training them with a smaller number of subjects, utilizing FL. We conducted two experiments involving varying numbers of subjects distributed across rural healthcare sites. In the first experiment, we allocated 10 subjects to site 1, 4 subjects to site 2, and 4 subjects to site 3. For the second experiment, we evenly distributed 6 subjects to each of the three sites.  For this experiment, we extracted 2D images from the MRI scans and allocated 80 percent for training and 20 percent for testing. In the context of performance assessment, we employed the Dice Similarity Coefficient (DSC), Sensitivity, Specificity, and Mean Absolute Error (MAE) as metrics. 

\begin{equation}
    DSC = \frac{2TP}{2TP+FP+FN}
\end{equation}

\begin{equation}
    Sensitivity = \frac{TP}{TP+FN}
\end{equation}

\begin{equation}
    Specificity = \frac{TN}{TN+FP}
\end{equation}

\begin{equation}
    MAE = \frac{\left| y - \hat{y}\right|}{n}
\end{equation}

where TP is true positive, FP is false positive, FN is false negative, TN is true negative, $y$ is actual value, and $\hat{y}$ is predicted value.
\subsection{Evaluation of local models at each site (without FL)}
First, we trained the local model using only the locally available data. This environment simulates a scenario where no information or parameter is shared with other healthcare sites, allowing us to evaluate the model's performance in isolation. At each rural site, the model is trained using local data. We employed the early stopping method to halt the training process at all sites. For both experiments with varying subject distribution, the evaluation matrix scores are presented in Table \ref{tab1}.

Table \ref{tab1} presents a comparative analysis of segmentation results obtained with the DRL model, indicating suboptimal performance. Specifically, in both Experiment 1 and Experiment 2, the DSC scores predominantly range from 70\% to 80\%, signifying the necessity for refinement. In Experiment 1, the DSC score for Site 1, featuring 10 subjects, exhibited a notable enhancement from 76\% to 94\%, a pattern observed across other sites as well. Meanwhile, in Experiment 2, where subject numbers were equitably distributed (6 subjects per rural site), DSC scores demonstrated marked improvement, increasing from an average of 77\% to as high as 93\%. Consequently, it is reasonable to deduce that the proposed framework, combining DRL and a refinement model, is well-suited for healthcare sites with a limited number of patients. This finding implies that rural healthcare facilities need not share patient data for further processing or diagnostic purposes utilizing segmentation models.

\subsection{Evaluation of local models trained with FL}
In the second training method, during the training stage of FL, initial local model training takes place. Subsequently, the parameters undergo transmission to the server and are aggregated. The aggregated results are then broadcasted back to the local healthcare sites. This iterative process continues until saturation is observed in the aggregated training error. Specifically for our training approach, prior to transmitting the parameters to the server, we conduct model training for 10 epochs in each iteration. Following the reception of aggregated parameters, the training for 10 epochs is reiterated, and error monitoring takes place after each iteration of the parameters' broadcast.

The outcomes of the proposed framework incorporating FL training are presented in Table \ref{tab2}. Notably, the performance of the DRL model following FL training exhibits minimal alteration, with segmentation DSC maintaining a consistent range between 70\% and 80\%, while Sensitivity, Specificity, and MAE exhibit uniformity across all sites for the DRL model alone. However, discernible enhancements emerge after FL training when the framework combines DRL with an RM.

In Experiment 1, the DSC for Site 2 experiences an advancement from 89\% to 92\%, and a similar improvement is observed for Site 3, elevating its DSC from 93\% to 95\%. Additionally, a reduction in MAE is observed for both sites. In Experiment 2, the introduction of FL training yields improved scores. A comparative analysis between the results of Experiment 2 in Table \ref{tab1} and Table \ref{tab2} demonstrates that DSC values for all three sites exhibit enhancement. Notably, Sites 1 and 2 surpass a 94\% DSC, reflecting a 4\% increase compared to training without FL. Consequently, it can be asserted that the proposed framework, when coupled with FL, exhibits superior performance. Notably, this approach ensures data privacy compliance by exclusively sharing model parameters during FL training while concurrently enhancing accuracy, even in scenarios with a limited number of subjects at rural sites.

The conclusive outcomes derived from the proposed framework are visually depicted in Fig. \ref{fig4}. In this figure, each input image is presented in the initial column, followed by the corresponding outputs generated by the DRL model, which are illustrated in columns 2, 3, and 4. It is noteworthy that the DRL environment undergoes three iterations for each input image, thereby yielding three coarsely segmented images for each input image. The ultimate refined segmented image, achieved through the refinement process, is showcased in the last column of Fig. \ref{fig4}.

\section{Conclusion}
In conclusion, our study presents an innovative segmentation framework tailored for federated learning applications within healthcare settings characterized by a restricted pool of subjects. This framework incorporates a DRL model capable of executing adaptive threshold-based segmentation, which is subsequently refined using RM to enhance the precision of coarse segmentation. This novel approach exhibits promise in segmentation with accuracy up to 92\% - 95\% for healthcare sites in rural areas that contend with limited data resources, all while preserving data privacy. Importantly, our model contributes to the reduction of computational complexity with a three times reduction in parameters than other segmentation models in the context of brain tissue segmentation without compromising the accuracy of the segmentation process. As we look toward the future, the ongoing development of this research will entail expanding our analysis to encompass larger data sets and incorporating multiple rural healthcare sites, employing various federated learning frameworks, and extending the scope to encompass the segmentation of all three tissue types in brain MRI scans.

\bibliographystyle{ieeetr}
\bibliography{RP_ICC2024}

\end{document}